\documentclass[letterpaper, 10 pt, conference]{ieeeconf}  
\usepackage{tabulary,graphicx,times,caption,fancyhdr,amsfonts,amssymb,amsbsy,latexsym,amsmath}
\usepackage{array}
\usepackage[caption=false,font=normalsize,labelfont=sf,textfont=sf]{subfig}
\usepackage{textcomp}
\usepackage{stfloats}
\usepackage{url}
\usepackage{verbatim}
\usepackage{graphicx}
\usepackage{cite}
\usepackage[utf8]{inputenc}
\usepackage{svg}
\usepackage{amssymb}
\usepackage{bbm}
\usepackage{hyperref}
\usepackage{relsize}%
\usepackage{graphicx}
\usepackage{subfig}
\usepackage{xcolor}
\usepackage[english]{babel}
\usepackage{xurl}

\usepackage{algorithmicx}
\usepackage{booktabs} 
\usepackage[ruled,vlined, linesnumbered]{algorithm2e}

\newcommand \mc \mathcal

\newcommand {\augm} [1] { \pmb{{#1}}}  

\newcommand {\x} {\augm x}

\IEEEoverridecommandlockouts                              

\title{\LARGE \bf
Delay-Aware Multi-Agent Reinforcement Learning for Cooperative Adaptive Cruise Control with Model-based Stability Enhancement
}

\author{Jiaqi Liu, Ziran Wang,~\IEEEmembership{Member,~IEEE,} Peng Hang,~\IEEEmembership{Member,~IEEE,} and Jian Sun
\thanks{This work was supported in part by the National Natural Science Foundation of China (52302502, 52232015), the Young Elite Scientists Sponsorship Program by CAST (2022QNRC001),  the State Key Laboratory of Intelligent Green Vehicle and Mobility under Project No. KFZ2408, the Belt and Road Cooperation Program under the 2023 Shanghai Action Plan for Science,Technology and Imnovation (23210750500), the Shanghai Scientific Innovation Foundation (No.23DZ1203400) and the Fundamental Research Funds for the Central Universities.}
\thanks{Jiaqi Liu, Peng Hang, and Jian Sun are with the Department of
Traffic Engineering and Key Laboratory of Road and Traffic Engineering,
Ministry of Education, Tongji University, Shanghai 201804, China. (e-mail: \{liujiaqi13, hangpeng,sunjian\}@tongji.edu.cn)}
\thanks{Ziran Wang is with College of Engineering, Purdue University,
West Lafayette, IN. (e-mail: ziran@purdue.edu)}
\thanks{Corresponding author: Peng Hang}
}

\begin{document}

\maketitle
\thispagestyle{empty}
\pagestyle{empty}

\begin{abstract}
Cooperative Adaptive Cruise Control (CACC) represents a quintessential control strategy for orchestrating vehicular platoon movement within Connected and Automated Vehicle (CAV) systems, significantly enhancing traffic efficiency and reducing energy consumption. 
In recent years, the data-driven methods, such as reinforcement learning (RL), have been employed to address this task due to their significant advantages in terms of efficiency and flexibility. However, the delay issue, which often arises in real-world CACC systems, is rarely taken into account by current RL-based approaches.
To tackle this problem, we propose a Delay-Aware Multi-Agent Reinforcement Learning (DAMARL) framework aimed at achieving safe and stable control for CACC. We model the entire decision-making process using a Multi-Agent Delay-Aware Markov Decision Process (MADA-MDP) and develop a centralized training with decentralized execution (CTDE) MARL framework for distributed control of CACC platoons. An attention mechanism-integrated policy network is introduced to enhance the performance of CAV communication and decision-making. Additionally, a velocity optimization model-based action filter is incorporated to further ensure the stability of the platoon. Experimental results across various delay conditions and platoon sizes demonstrate that our approach consistently outperforms baseline methods in terms of platoon safety, stability and overall performance.
\end{abstract}

\section{Introduction}
Connected and Automated Vehicles (CAVs) are poised to bring revolutionary changes to the entire transportation system. Cooperative Adaptive Cruise Control (CACC) represents a pivotal technology within CAVs, where each CAV is required to formulate an independent driving plan\cite{lesch2021overview}. Through coordination with other CAVs, the technology aims to avoid potential collisions and achieve the objective of string stability in CACC platoons. By adaptively coordinating the platoon through vehicle-to-vehicle communication, CACC can significantly reduce the following distances and speed variations while maintaining safety, thereby enhancing traffic efficiency, alleviating congestion, and reducing energy consumption\cite{lesch2021overview,wang2017developing,wang2018review}.

In recent years, CACC technology has garnered considerable attention from researchers, leading to extensive exploration. Common methodologies include classical control theory and optimization-based approaches \cite{massera2017safe, wu2018stabilizing}.  
Specifically, some studies have focused on car-following models \cite{massera2017safe} and string stability \cite{wang2018infrastructure, feng2019string}, modeling CACC within the context of two-vehicle systems.
Other researchers have conceptualized CACC as an optimal control problem \cite{gao2016data, wu2018stabilizing}. These methods rely on precise system modeling, which is often unavailable \cite{wang2018infrastructure,feng2019string}.

On the other hand, CACC platoon control has also been conceptualized as a sequential decision-making problem and modeled using reinforcement learning (RL) \cite{peake2020multi, chu2019model,lei2022deep, jiang2022reinforcement, liu2022autonomous}.
RL has experienced rapid advancements in recent years \cite{liu2023mtd, liu2023towards}, originally proposed in the control domain within the Markov Decision Process (MDP) framework for optimal stochastic control under uncertainty \cite{sutton2018reinforcement}. 
This approach demonstrates remarkable efficiency and flexibility, making it a promising candidate for CACC platoon control\cite{chu2019model,lei2022deep}.

Peake et al. \cite{peake2020multi} have pioneered the application of deep reinforcement learning for CACC, enabling platoon vehicles to adopt a robust communication protocol and facilitating vehicle training through LSTM-based policy networks. Concurrently, Wang et al. explored the utilization of the policy iteration method to deduce parameters for the classical Proportional-Integral (PI) controller, sidestepping the need for direct longitudinal control \cite{wang2013self}. Further, a novel information reward design was proposed to bolster the safety and robustness of the Q-learning technique \cite{li2017training}, while \cite{desjardins2011cooperative} introduced a policy gradient RL strategy aimed at preserving safe longitudinal distances between vehicles. 

However, it is noteworthy that, despite the significant attention and research on RL methods for CACC control problems, these approaches still rely on rather idealized assumptions in their modeling, such as perfect or delay-free communication and decision-making. This impedes the further deployment and application of these methods in the real world. In reality, communication delays within CACC, sensor latency, and delays in the execution of decision actions are common and inevitable\cite{wang2022design}. Existing RL methods rarely consider these time-delay issues. Particularly for multi-agent problems like CACC, delays can exacerbate issues, as a delay in one agent may propagate to other coupled agents. This not only can lead to a decline in agent performance but also may disrupt the stability of the dynamic system, potentially causing catastrophic failures in safety-critical systems\cite{chen2020delay}.

To address these challenges, we propose a Multi-Agent Reinforcement Learning (MARL) framework that incorporates delay awareness and an enhanced attention mechanism to model the collaborative decision-making problem in CACC systems. Initially, we consider the multiple delays present within the CACC platoon and construct a Multi-Agent Delay-Aware  Markov Decision Process (MADA-MDP) to model this problem. Subsequently, based on the MADA-MDP framework, we employ a MARL approach characterized by Centralized Training with Decentralized Execution (CTDE) to model the entire CACC decision-making process. In this framework, each CAV is controlled by an independent policy network, with complete information sharing during centralized training, while allowing for independent decision-making by each CAV beyond necessary platoon communication to enhance platoon safety. 
The attention mechanism is employed in the policy network to enhance the information processing ability of CAVs. Furthermore, to address the complexities of platoon decision-making, we introduce an action filter based on a velocity optimization model to the MARL's decision output, ensuring platoon stability.

We validated our approach on CACC platoons of various sizes. Compared with several baseline methods, our method effectively handles internal platoon delay issues and demonstrates superior performance in terms of platoon safety and stability.

Our contributions can be summarized as follows:
\begin{itemize}
    \item We propose a Multi-Agent Markov decision process that considers the internal delays within CACC, and based on this process, we formulate the CACC  problem within a CTDE MARL framework.
    \item We incorporate an attention mechanism-based decision-making policy network for CAVs and design a model-based action filter for each CAV to enhance the safety and stability of the platoon.
    \item Our method is tested on fleets of various sizes within simulation environments, and compared to benchmark algorithms, it demonstrates outstanding performance.
\end{itemize}

\begin{figure}[!htbp]
    \centering
    \includegraphics[width=1\linewidth]{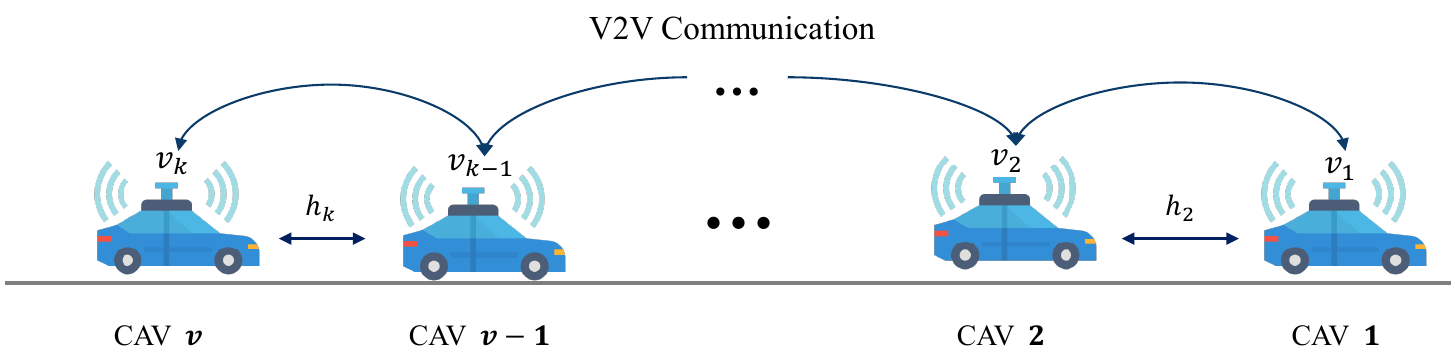}
    \caption{The overview of the CACC system.}
    \label{fig:platoon_overview}
\end{figure}



\section{Preliminary}  
\label{sec:2}
In this section, Markov Decision Processes(MDPs) under delay conditions and multi-agent scenarios are introduced.

\subsection{Delay-Aware Markov Decision Process}
\begin{figure}
    \centering
    \includegraphics[width=0.70\linewidth]{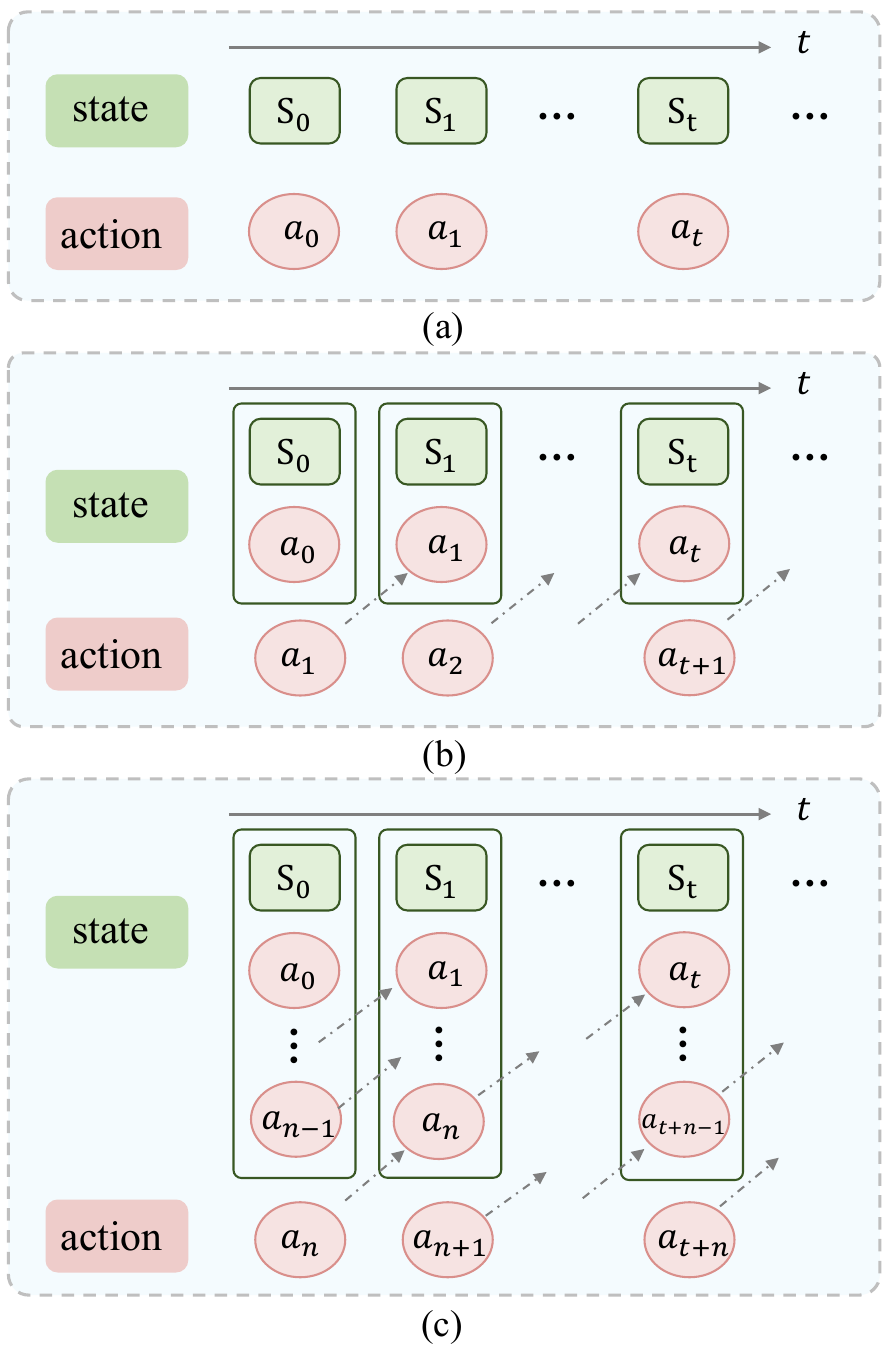}
    \caption{Comparison between (a) MDP$(E)$, (b) DMDP$(E, 1)$ and (c) DMDP$(E, n)$. $n$ denotes the action delay step. 
    }
    \label{fig:DAMDP}
\end{figure}
For decision systems where delay is not a consideration, the traditional Markov process (delay-free) can be employed for modeling. A traditional MDP is typically defined by a quintuple: $\text{MDP} = \langle S, A, \rho, p, r \rangle$\cite{liu2023optimal}. Here, $S$ denotes the state space, $A$ is the action space, $\rho$ represents the initial state distribution, $p$ is the transition distribution, and $r$ is the reward function. In RL, the MDP framework is often utilized to model the decision-making problem, with $\pi$ representing the policy of an agent. The objective of RL is to find the optimal policy $\pi^*$ that maximizes the expected cumulative reward $\sum_{t=0}^{T} \gamma^t r(s_t, a_t)$.
However, the traditional MDP framework's assumption of immediate action may lead to policies far from optimal in decision systems with delays, potentially causing significant performance degradation or safety issues.

To mitigate this problem and restore the Markov property, which means future states depend only on the current state and action), we introduce the Delay-Aware Markov Decision Process (DAMDP) to accommodate delays\cite{schuitema2010control,chen2020delay}.
The DAMDP represents an augmented variant of the traditional MDP, formally defined by the tuple $\langle \boldsymbol{\mathcal{X}}, \boldsymbol{\mathcal{A}}, \tilde{\rho}, \tilde{p}, \tilde{r} \rangle$. Here, $\boldsymbol{\mathcal{X}} = \mathcal{S} \times \mathcal{A}^{k}$ encapsulates the state space, augmented to incorporate the action sequence to be executed over the next $k$ steps, with $k \in \mathbb{N}$ representing the delay duration. The action space $\boldsymbol{\mathcal{A}}$ remains consistent with that of the original MDP. The augmented reward function is defined as $\tilde{r}(\mathbf{x}_t, \tilde{a}_t) = r(s_t, a_t)$, and the transition distribution is given by:
\begin{equation}
    \begin{aligned}
    \tilde{p}(\mathbf{x}_{t+1}|\mathbf{x}_t, \tilde{a}_t) = & p(s_{t+1} | s_t, a_t^{(t)}) \prod_{i=1}^{k-1}\delta(a_{t+i}^{(t+1)} - a_{t+i}^{(t)}) \cdot \\
    & \delta(a_{t+k}^{(t+1)} - \tilde{a}_t),
    \end{aligned}
\end{equation}
where the state vector $\mathbf{x}$ of DAMDP includes an action sequence to be executed in the forthcoming $k$ steps. The notation $a_{t_1}^{(t_2)}$ denotes that the action is an element of $\mathbf{x}_{t_2}$, with the subscript indicating the time of action execution, and the superscript specifying the reference time frame. $\delta$ is the Dirac delta function. If $y \sim \delta(\cdot - x)$ then $y=x$ with probability one.

The action $\tilde{a}_t$, selected at time $t$ within a DAMDP context, is slated for execution at time $t+k$, thereby accommodating the $k$-step action delay, such that $\tilde{a}_t= a_{t+k}$. Furthermore, the policy network $\pi$ for the agent is under continuous refinement to ascertain the optimal policy, adjusting dynamically to the intricacies introduced by the delay-aware framework.

\subsection{Multi-Agent Delay-Aware Markov Decision Process}
Given the prevalence of various types of delays in real-world decision systems, we define the Multi-Agent Delay-Aware Markov Decision Process (MADA-MDP) based on DAMDP framework for subsequent modeling of CACC vehicle platoons. The MADA-MDP is articulated as the tuple $\langle \boldsymbol{\mathcal{X}}, \boldsymbol{\mathcal{A}}, \rho, p, r \rangle$, where $\boldsymbol{\mathcal{X}} = \mathcal{S} \times \mathcal{A}_{1}^{k_1} \times \dots \times \mathcal{A}_{N}^{k_N}$ represents the augmented state space, incorporating the action sequence to be executed over the next $k_i$ steps for agent $i$, thereby denoting the delay step. The action space $\boldsymbol{\mathcal{A}}$ aligns with that in the traditional MDP.

The transition distribution is defined as:
\begin{equation}
\begin{aligned}
& p(\mathbf{x}_{t+1}|\mathbf{x}_t, \tilde{a}_t) = \\
& p\left(s_{t+1}, a^{1, (t+1)}_{t+1}, \dots, a^{1, (t+1)}_{t+k_1},\dots, a^{N, (t+1)}_{t+1}, \dots, a^{N, (t+1)}_{t+k_N} \right. \\
& \left. | s_t, a^{1, (t)}_{t}, \dots, a^{1, (t)}_{t+k_1-1},\dots, a^{N, (t)}_{t}, \dots, a^{N, (t)}_{t+k_N-1}, \tilde{a}_t \right) \\
= & p(s_{t+1} | s_t, a^{1, (t)}_t, \dots, a^{N, (t)}_t) \\
& \prod_{i=1}^{N}\prod_{j=1}^{k_i-1}\delta(a^{i, (t+1)}_{t+j} - a^{i, (t)}_{t+j}) \prod_{i=1}^{N} \delta(a^{i, (t+1)}_{t+k_i} - \tilde{a}_t^i),
\end{aligned}
\end{equation}
where the policy input for agent $i$ at time $t$ comprises two components: $\tilde{o}_t^i = (o_{t, \text{obs}}^i, o_{t, \text{act}}^i)$, with $o_{t, \text{obs}}^i$ being the observation of the environment and $o_{t, \text{act}}^i$ representing a planned action sequence of length $k_i$ for agent $i$, to be executed from the current time step: $o_{t, \text{act}}^i = (a_t^i, \dots, a_{t+k_i-1}^i)$.

Agents $i$ take actions $a_i \in \boldsymbol{\mathcal{A}_i}$ based on a policy $\pi_i : \boldsymbol{\mathcal{X}_i} \times \boldsymbol{\mathcal{A}_i} \to [0,1]$, interacting with the MADA-MDP environment, which is augmented due to the extended dimensionality of state vectors resulting from action and observation delays. Notably, action and observation delays constitute similar mathematical challenges, as they both introduce a mismatch between current observations and executed actions, as proven in literature \cite{katsikopoulos2003markov}. 
To ensure conciseness in our study, we have considered and transformed both the observation delay and action delay into a mathematical formula representing action delays.

\section{Problem Formulation for CACC}
\label{sec:3}
In this section, we introduce the vehicle dynamics model utilized for modeling CACC systems and formalize the CACC problem using the MADA-MDP.

\subsection{Vehicle Dynamics}
\label{subsec:dynamics}
As depicted in Fig.\ref{fig:platoon_overview}, we consider a platoon comprising $\mathcal{V}$ CAVs traveling along a straight path. For the sake of simplicity, it is assumed that all CAV in the system possess identical characteristics, such as maximum permissible acceleration and deceleration. The platooning system is orchestrated by a platoon leader vehicle (PL, 1\textsuperscript{st} vehicle), while the platoon member vehicles (PMs, $i \in \{2, \ldots, \mathcal{V}\}$) follow the PL. Each PM $i$ maintains a desired inter-vehicle distance (IVD) $h_i$ and velocity $v_i$ relative to its preceding vehicle $i-1$, adhering to its distinct spacing policy \cite{zhu2022joint}. The one-dimensional dynamics for vehicle $i$ are given by:
\begin{subequations}
\begin{IEEEeqnarray}{rCl}
\dot{h}_i &=& v_{i-1} - v_i,\\
\dot{v}_i &=& u_i,
\end{IEEEeqnarray}  
\end{subequations}
where $v_{i-1}$ and $u_i$ denote the velocity of the preceding vehicle and the acceleration of vehicle $i$, respectively. Following the framework established in \cite{chu2019model}, the discretized vehicle dynamics, with a sampling interval $\Delta t$, are described by:
\begin{subequations}
\begin{IEEEeqnarray}{rCl}
h_{i, t+1} &=& h_{i, t} + \int_{t}^{t + \Delta t} (v_{i-1, \tau} - v_{i, \tau}) d\tau,\\
v_{i, t+1} &=& v_{i, t} + u_{i, t} \Delta t.
\end{IEEEeqnarray}  
\end{subequations}

To ensure both comfort and safety, each vehicle must adhere to the following constraints \cite{chu2019model}:
\begin{subequations}
\begin{IEEEeqnarray}{rCl}
h_{i, t} &\geq& h_{\text{min}},\\
0 &\leq& v_{i,t} \leq v_{\text{max}},\\
u_{\text{min}} &\leq& u_{i,t} \leq u_{\text{max}},
\end{IEEEeqnarray}  
\end{subequations}
where $h_{\text{min}}$, $v_{\text{max}}$, $u_{\text{min}}$, and $u_{\text{max}}$ represent the minimum safe headway, maximum speed, deceleration, and acceleration limits, respectively.

\subsection{Problem Formulation}
Considering the prevalent issue of delays stemming from communication or decision-making processes within CACC systems in the real world, we model this problem using the MADA-MDP. The process is represented by the tuple $\langle \boldsymbol{\mathcal{X}}, \boldsymbol{\mathcal{A}},  \rho, p, r \rangle$, where $\boldsymbol{\mathcal{X}}$ denotes the augmented state space, $\mathcal{A}$ represents the action space, and the state transition distribution $p$ characterizes the system's intrinsic dynamics.
Furthermore, we assume that each CAV is equipped with the capability to communicate with the vehicles ahead and behind via V2V communication channels.

\subsubsection{State Space}
The state space encapsulates the environmental description. Following the definition of MADA-MDP, the observation spaces of all CAVs are augmented as $\boldsymbol{\mathcal{X}} = \mathcal{S} \times \mathcal{A}_{1}^{k_1} \times \dots \times \mathcal{A}_{N}^{k_N}$. Here, the original state of agent $i$, $\mathcal{S}_i$, is defined by $[v, v_{\text{diff}}, v_h, h, u]$, where $v$ denotes the current normalized vehicle speed. $v_{\text{diff}} = (v_{{i-1},t} - v_{i,t})$ represents the vehicle speed difference with its leading vehicle. $v_h = (v^\circ(h) - v_{i,t})$, $h = (h_{i,t} + (v_{i-1, t} - v_{i,t}) \Delta t - h^*) / h^*$, and $u = u_{i, t} / u_{\text{max}}$ are the headway-based velocity, normalized headway distance, and acceleration, respectively.

\subsubsection{Action Space}
In the considered CACC system, the action $a_t \in \mathcal{A}$ directly pertains to the longitudinal control. The action decision output for CAV $i$ includes three components $a_{i,t}=(\alpha_{i,t},\beta_{i,t},\hat{u}_{i,t})$, where $\hat{u}_{i,t}$ is the acceleration control value output by the policy network, and $\alpha_{i,t},\beta_{i,t}$ are control parameters for the action filtering layer, which will be elaborated upon in Section \ref{sec:speed_controller}. The overall action space comprises the combined actions of all CAVs, i.e., $\boldsymbol{\mathcal{A}} = \mathcal{A}_{1} \times \mathcal{A}_{2} \times \cdots \times \mathcal{A}_{|\mathcal{V}|}$.

\subsection{Reward Function}
To ensure that CAVs within the platoon simultaneously prioritize safety, efficiency, and comfort, our reward function is designed as follows:
\begin{equation} 
\label{eqn:reward_fn}
R_{i, t} = \frac{1}{C} \big[ w_1 (h_{i, t} - h^*)^2 + w_2 (v_{i, t} - v^*)^2 + w_3 u^2_{i, t} \big],
\end{equation}
where $w_1$, $w_2$, and $w_3$ are weighting coefficients, and $C$ is the scaling coefficient. In this equation, the first two terms, $(h_{i, t} - h^*)^2$ and $(v_{i, t} - v^*)^2$, penalize deviations from the desired headway and speed, respectively, encouraging the agent to closely achieve these objectives. The third term, $u^2_{i, t}$, is included to minimize sudden accelerations, thereby promoting a smoother and more comfortable ride for passengers.

\section{Delay-aware MARL for CACC Platoons}
\label{sec:4}
\begin{figure*}
    \centering
    \includegraphics[width=0.70\linewidth]{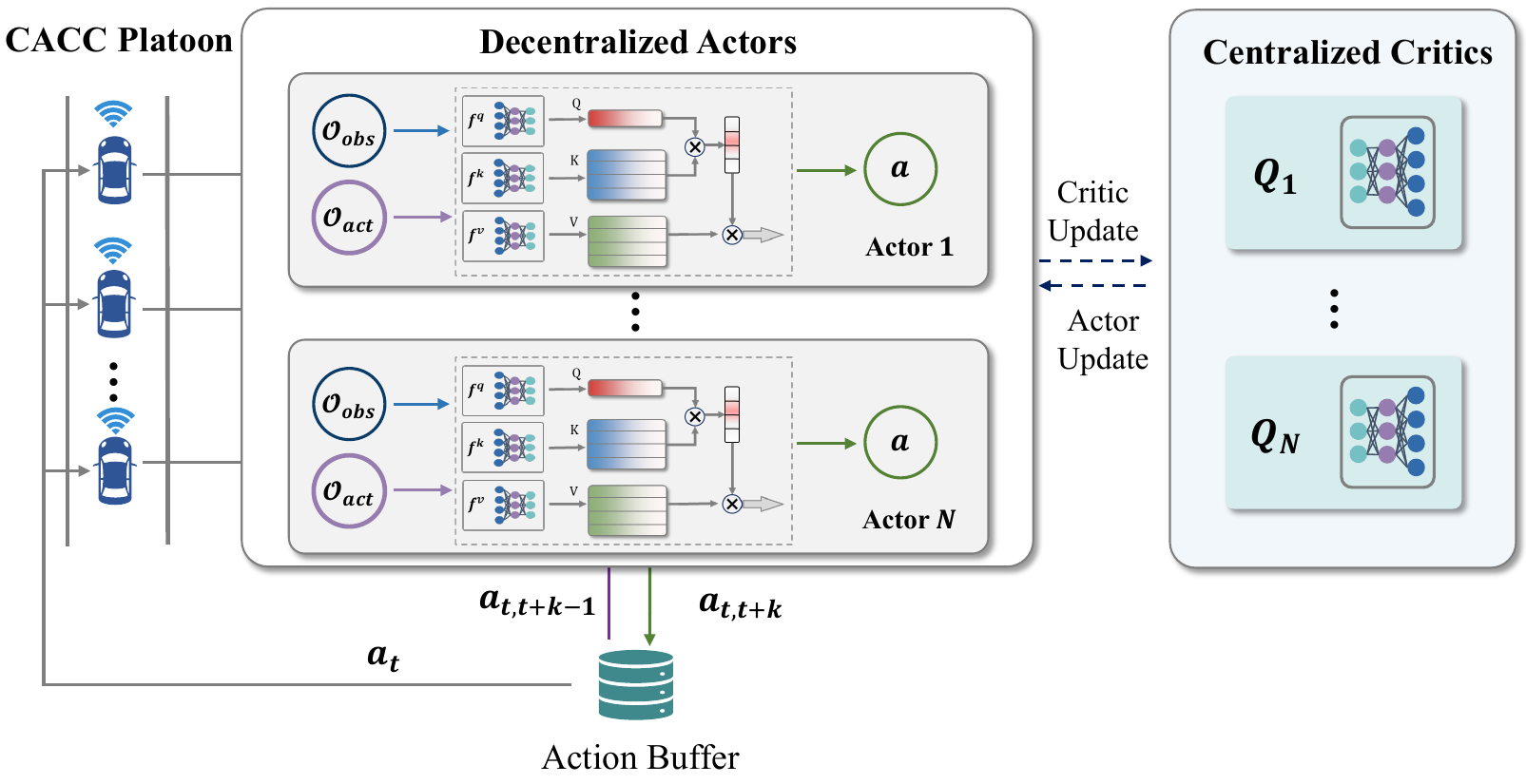}
    \caption{The framework of DAMARL for CACC platoon.}
    \label{fig:DAMARL_for_CACC}
\end{figure*}

In this section, we first introduce the framework of Delay-Aware MARL. Subsequently, we provide a detailed description of the policy network and the stability enhancement module.


\subsection{Delay-Aware Multi Agent Reinforcement Learning}
Drawing from the findings of\cite{chen2020delay}, it is understood that MDPs under delay conditions do not necessitate direct modeling but can instead be directly solved through RL. Hence, based on MADA-MDP, we propose the Delay-Aware Multi-Agent Reinforcement Learning-CACC (DAMARL-CACC) decision framework. This approach employs a centralized training and decentralized execution paradigm, meaning that the vehicle platoon does not require a central controller, effectively mitigating the non-stationarity problem posed by multi-agent decision-making.

For each CAV in the platoon, we introduce a centralized Q-function Critic, based on global information, and an independent Actor that only requires communication and partial observations. The framework is illustrated in Fig.\ref{fig:DAMARL_for_CACC}. In this framework, each CAV $i$ is controlled by an independent policy network $\pi_i$, with $\pi=\{\pi_1,\ldots,\pi_N\}$ representing the set of all policies, parameterized by $\theta=\{\theta_1,\ldots,\theta_N\}$. The objective function's gradient for CAV $i$ is updated using the deterministic policy gradient algorithm as follows:


\begin{equation}
\label{equ:dadpg}
\begin{aligned}
& \nabla_{\theta_i} J(\mu_i) =  \\
&\mathbb{E}_{x, a \sim \mathcal{D}}[\nabla_{\theta_i} \mu_i(a_i| o_i) \nabla_{a_i} Q^{\mu}_i (x, a_1, ..., a_N)|_{a_i= \mu_i (o_i)}],
\end{aligned}
\end{equation}
where policies $\pi$, states $x$, and observations $o$ are augmented based on the MADA-MDP. In the delay-aware scenario, $\boldsymbol{\mathcal{X}}$ includes all CAVs' observations and the action sequences of all agents in the near future, $\boldsymbol{\mathcal{X}} = (o_1, ..., o_N)$, where $o_i$ is the input to CAV $i$'s policy $\pi_i$ and consists of two parts: $o_i = (o_{obs}^i, o_{act}^i)$. Here, $o_{obs}^i$ is the $i^{th}$ CAV's observation of the environment, and $o_{act}^i$ is the planned action sequence of length $k_i$ for agent $i$ starting from the current timestep, e.g., at time $t$, $o_{t, act}^i = a_{t:t+k_i-1}^i$. $o_{act}^i$ is obtained from the action buffer, serving as a bridge between the agent and the environment.

The replay buffer $\mathcal{D}$ is used to record the historical experiences of all CAVs. The centralized Q function $Q^{\mu}_i$ for CAV $i$ is updated as\cite{liu2023cooperative}:

\begin{equation*}
\mathcal{L}(\theta_i) = \mathbb{E}_{x, a, r, x'}\left[(Q^{\mu}_i(x, a_1,\ldots, a_N) - y)^2\right], \quad 
\end{equation*}
where $y = r_i + \gamma Q^{\mu'}_i(x', a_1',\ldots, a_N')\big|_{a_j'= \mu'_j(o_j)}$, and $\mu' = \{\mu_{\theta'_1}, \ldots, \mu_{\theta'_N}\}$ representing the target policy set with softly updated parameters $\theta'_i$ to stabilize training.

\subsection{Attention-based Policy Network}
\begin{figure*}
    \centering
    \includegraphics[width=0.8\linewidth]{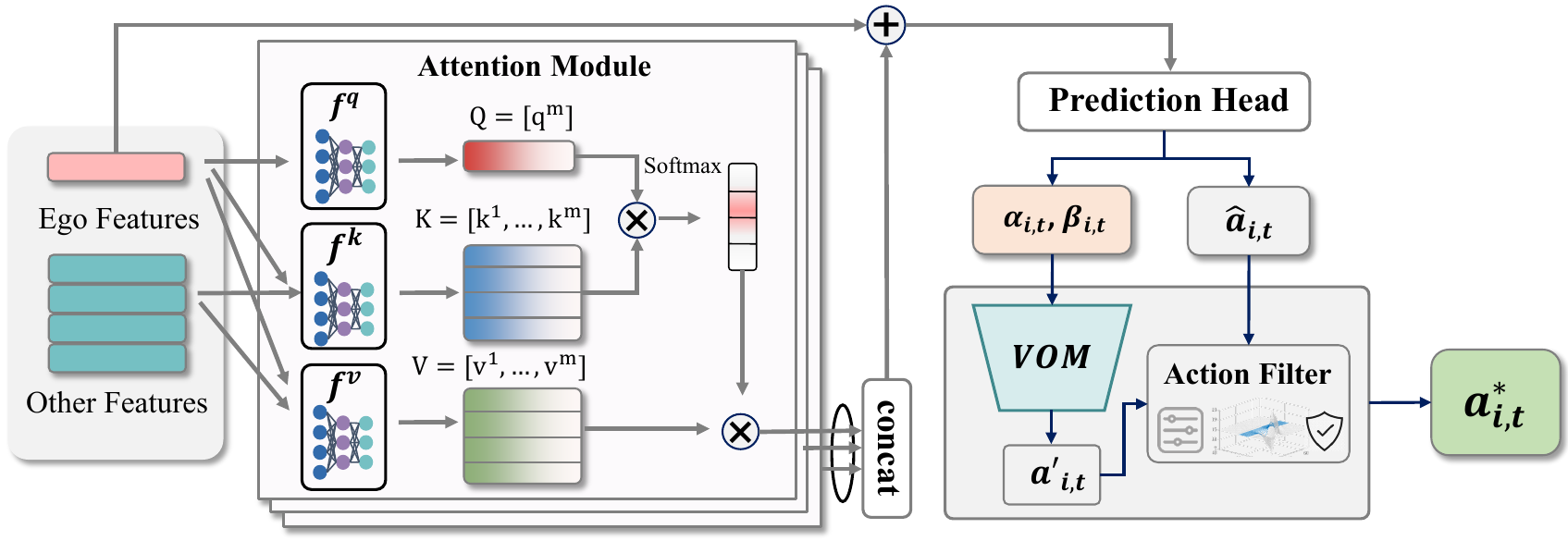}
    \caption{The attention-based policy network with stability enhancement.}
    \label{fig:policy_safety}
\end{figure*}

Recent advancements have underscored the efficacy of attention mechanisms in enabling neural networks to prioritize the processing of the most pertinent input information, thereby augmenting the model's computational efficiency \cite{liu2023towards,liu2023mtd}. In light of these findings, we have developed an attention-based policy network for each CAV, as delineated in Fig. \ref{fig:policy_safety}. This network is structured around three core components: the encoder block, the attention block, and the decoder block.

\textbf{Encoder Block:} The features $\mathcal{X}$ of agent $i$ are transformed into high-dimensional vectors through the application of a Multilayer Perceptron (MLP), with a uniform set of weights shared across all vehicles:
\begin{equation}
\label{eq:state_encoding}
    \mathcal{Z}_i = MLP(\mathcal{X}_i)
\end{equation}
Subsequently, the feature matrix is channeled into the attention block, which is composed of $N_{head}$ stacked attention heads.

\textbf{Attention Block:} Each self-vehicle generates a singular query $Q_i=[q_0] \in \mathbbm{R} ^{1\times d_k}$, aimed at identifying a subset of vehicles based on the environmental context, where $d_k$ denotes the output dimensionality of the encoder layer. This query is subjected to a linear projection and juxtaposed against a set of keys $K_i=[k^0_i,k^1_i, \ldots,k^N_i]\in \mathbbm{R}^{(N+1)\times d_k}$, each encapsulating descriptive features of individual vehicles. The similarity is computed using the dot product $q_0 k_i^T$. The components $Q_i$, $K_i$, and $V_i$ are determined as follows:
\begin{equation}
\label{eq:attention_matrix_mapping}
    \begin{aligned}
        & Q_i = W^Q \mathcal{Z}_i \\
        & K_i = W^K \mathcal{Z}_i \\
        & V_i = W^V \mathcal{Z}_i
    \end{aligned}
\end{equation}
where the dimensions for $W^Q$ and $W^K$ are configured as $(d_k\times d_N)$, and that for $W^V$ is $(d_v \times d_h)$.
The computation of the attention matrix involves scaling the dot product by the inverse square root of the dimension $\frac{1}{\sqrt{d_k}}$ and its normalization using a softmax function $\sigma$. This process facilitates the aggregation of a set of output values $V_i=[v^0_i,\ldots,v^N_i]$, wherein each $v^j_i$ represents a feature derived through a shared linear projection $L_v \in \mathbbm{R}^{d_x \times d_k}$. The attention calculation for each head is encapsulated by:
\begin{equation}
\label{eq:cal_attention_1}
    \omega_i^m = \sigma \Big (  \frac{Q_i K^T_i}{\sqrt{d_k}} \Big)V_i
\end{equation}

The aggregated output from all $M$ heads is unified through a linear layer:
\begin{equation}
\label{eq:cal_attention_2}
    \omega_i = \sum_{m=1}^{M} \omega_i^m
\end{equation}

The resultant attention vector $\omega_i=\big[\omega_{i,1}, \omega_{i,2},\ldots, \omega_{i,|\mathcal{N}_i|} \big]$ quantifies the degree of attention agent $i$ allocates to the surrounding vehicle $j$, conforming to the cumulative summation condition: $\sum^{|\mathcal{N}_i|}_{j=1} \omega_{i,j} = 1$.

\textbf{Decoder Block:} Leveraging the attention-weighted features $\omega_i$ alongside the encoded state information $\mathcal{Z}_i$, the decoder employs a MLP to deduce the final action vector:
\begin{equation}
\label{eq:final_action_vector}
    a_i = MLP(\mathcal{Z}_i, \omega_i)
\end{equation}
Then the action $a_i$ is checked by an action filter to ensure the security and stability, which is described in the next subsection.

\subsection{Platoon Stability Enhancement with Model-based Controller}
\label{sec:speed_controller}
Given MARL's data-driven essence, we introduce an action filter layer comprising a velocity optimization model to enhance the safety and robustness of CACC platoons by ensuring safe acceleration for trailing CAVs.

The velocity optimization model computes a safe acceleration, termed the ideal acceleration $u'_{i,t}$ for the $i^{th}$ CAV, based on control parameters:
\begin{equation} \label{eqn:ovm1}
u'_{i,t} = \alpha_i (v^\circ (h_{i, t}; h^s, h^g) - v_{i, t}) + \beta_i (v_{i-1, t} - v_{i,t}),
\end{equation}
where $\alpha_i$ and $\beta_i$ denote the headway gain and relative speed gain, respectively, influencing vehicle acceleration. Here, $h^s$ and $h^g$ symbolize the stopped and full-speed headway distances.

The headway-based velocity strategy $v^\circ$ is defined as:
\begin{equation} \label{eqn:ovm2}
v^\circ (h) \triangleq
\begin{cases}
0, & \text{if } h < h^s,\\
\frac{1}{2} v_{max} (1 - \cos{(\pi \frac{h - h^s}{h^g - h^s})}), & \text{if } h^s \leq h \leq h^g,\\
v_{max}, & \text{if } h > h^g.
\end{cases}
\end{equation}
This function outlines an optimal velocity based on the current headway, with velocity set to zero for headways $\leq h^s$ to prevent collisions, gradually increases within $h^s$ to $h^g$, and is capped at $v_{max}$ for headways $\geq h^g$, ensuring traffic flow safety and efficiency.

The action output by each CAV at time $t$, $a_{i,t} = (\alpha_{i,t}, \beta_{i,t}, \hat{u}_{i,t})$, comprises the CAV's acceleration control value $\hat{u}_{i,t}$, determined by the policy network, and $\alpha_{i,t}, \beta_{i,t}$, the control parameters for the action filter layer.
Upon action generation by the policy network, the action filter layer computes the ideal acceleration $u'_{i,t}$ using parameters $\alpha_{i,t}, \beta_{i,t}$ and Equation \eqref{eqn:ovm1}. It then selects the acceleration action $u_{i,t}^*$ that maximizes benefits according to the reward function:
\begin{equation}
\label{eq:action_filter}
    u_{i,t}^* = \begin{cases} 
    u'_{i,t} & \text{if } R(u'_{i,t}) \geq R(\hat{u}_{i,t}), \\
    \hat{u}_{i,t} & \text{otherwise},
    \end{cases}
\end{equation}
where $R(\cdot)$ denotes the reward function specified in Equation \eqref{eqn:reward_fn}.

For a detailed description of the entire algorithm, see Algorithm\ref{alg:damarl}.

\begin{algorithm}
\caption{DAMARL for CACC}
\label{alg:damarl}
\resizebox{\columnwidth}{!}{
\begin{minipage}{\columnwidth}

\SetAlFnt{\small}
\SetKwInOut{Parameter}{Inputs}
\SetKwInOut{Output}{Outputs}
\LinesNumbered 
\SetAlgoLined
\Parameter{Replay buffer $\mathcal{D}$, Episodes $M$, Timesteps $T$, CAVs $N$, Batch size $B$}
\Output{Updated policies $\theta_i'$ and centralized critics for each CAV $i$}
\vspace{0.2em}
\hrule
\vspace{0.2em}
Initialize the experience replay buffer $\mathcal{D}$;\\
\For{$\textrm{episode}=1$ \KwTo $M$}
{
    Initialize the action noise $\mathcal{N}_t$ and the action buffer $\mathcal{F}$;\\
    Get initial state $\x_0$;\\
    \For{$t=1$ \KwTo $T$}
    {
        \For{CAV $i=1$ \KwTo $N$}
        {
            Obtain $ o_i = (o^i_{obs}, o^i_{act})$ from the environment and $\mathcal{F}$;\\
            \textit{\bfseries  Output action with stability check:}\\
            Get $(\hat{u}_{i}, \alpha_i, \beta_i)$ from $\pi_{\theta_i}(o_i)$;\\
            Calculate the ideal action by Equation \eqref{eqn:ovm1};\\
            Select optimal action $ a^*_i$ by Equation\eqref{eq:action_filter};\\
        }
        Store actions $ \textbf{a}=( a_1,\dots, a_N)$ in $\mathcal{F}$;\\
        Pop $ \textbf{a}=( a_1,\dots, a_N)$ from $\mathcal{F}$ and execute it;\\
        get the reward $ r$ and the new state $ x'$;\\
        Store $(\mathcal{\x}, \textbf{a}, r,\x') \rightarrow \mathcal{D}$;\\
        $\x \gets \x'$;\\
        \For{CAV $i=1$ \KwTo $N$}
        {
            Randomly sample a batch of $B$ samples $(\x^b, \textbf{a}^b, r^b, \x'^b)$ from $\mathcal{D}$;\\
            Set $y^b=r_i^b+\gamma\, Q_i^{ \pi'}(\x'^b, a_1',\dots, a_N')|_{ a_l'= \pi'_l( o_l^b)}$;\\
            Update centralized critics with loss $\mathcal{L}(\theta_i)=\frac{1}{B}\sum_j\left(y^b-Q_i^{ \pi}(\x^b, a^b_1,\dots,  a^b_N)\right)^2$;\\
            Update decentralized actors by $\nabla_{\theta_i}J \approx \frac{1}{B}\sum_j\nabla_{\theta_i} \pi_i( o_i^b)$ \\
            $\nabla_{a_i}Q_i^{ \pi}(\x^b, a^b_1,\dots,  a^b_N)\big|_{ a_i= \pi_i( o_i^b)}$;\\
        }
        Soft update of target networks for each agent $i$: $\theta_i'\gets\kappa\theta_i+(1-\kappa)\theta_i'$;\\
    }
}
\end{minipage}
}
\end{algorithm}

\section{Experiments and Analysis}
\label{sec:5}
\subsection{Simulation Settings}
\textbf{Scenarios.} 
In our study, we devise two distinct scenarios, namely Catchup and Slowdown, to evaluate the efficacy of our methods. In the Catchup scenario, PMs indexed by $i = 2, \ldots, \mathcal{V}$ are initialized with velocities $v_{i, 0} = v^*_t$ and headways $h_{i, 0} = h^*_t$. Conversely, the PL begins with $v_{1, 0}= v^*_t$ and $h_{1, 0} = a \cdot h^*_t$, where $a$ represents a random variable uniformly distributed between 3 and 4. The Slowdown scenario, on the other hand, initializes all vehicles, indexed by $i = 1, \ldots, \mathcal{V}$, with initial velocities $v_{i, 0} = b \cdot v^*_t$ and headways $h_{i, 0} = h^*_t$, with $b$ uniformly distributed between 1.5 and 2.5. Herein, $v^*_t$ linearly decreases to 15 m/s within the first 30 seconds and subsequently remains constant. For the dynamics model of CACC in subsection \ref{subsec:dynamics}, we set $h_{\text{min}} = 1$ m, $v_{\text{max}}=30$ m/s, $u_{\text{min}} = -2.0$ m/s  $\textsuperscript{2}$, and $u_{\text{max}} = 2.0$ m/s $\textsuperscript{2}$.

\textbf{Delay Setting.}  
In the real world, the predominant delays encountered in CACC encompass communication delay $\tau_1$, sensor delay $\tau_2$, and actuator delay $\tau_3$. We approximate the total delay $\tau$ as the cumulative sum of these components: $\tau=\tau_1+\tau_2+\tau_3$. Based on settings from the literature regarding communication and sensor delays\cite{xing2019compensation,liu2023communication}, we configure the total delay $\tau$ to be $0.5$ seconds.

\textbf{Baselines.}
For comparative analysis, we select three cutting-edge MARL algorithms as baselines, including MADDPG\cite{lowe2017multi}, ConseNet\cite{zhang2018fully}, and NeurComm\cite{chu2020multiagent}. MADDPG is a typical example of the CTDE approach used in multi-agent decision-making. ConseNet and NeurComm specifically incorporate designed communication protocols to facilitate enhanced decision-making among agents.

\subsection{Implementation Details}
We set the hyperparameters $w_1$, $w_{2}$, and $w_{3}$ in Equation~\eqref{eqn:reward_fn} to -1, -1, and -0.2, respectively, with $C=15$. The simulation spans $T=60$ seconds with an interaction period $\Delta t=0.1$ seconds, allowing the environment to progress $\Delta t$ seconds per MDP step. And the delay planned action sequence length for each CAV $i$ is calculated as: $k_i =\lfloor \frac{\tau}{\Delta t} \rfloor=5$.
The Encoder and Decoder in the attention-based policy network are MLP with two $64 \times 64$ linear layers. The Attention Layer has two heads and a 128 feature size. For the model-based controller, $h^s = 5$m and $h^g = 35$m.
We train over $1\text{M}$ steps for each model, with the discount factor $\gamma$ of $0.99$. The learning rates are set to $5.0 \times 10^{-4}$ for the actor and $2.5 \times 10^{-4}$ for the critic.
Each algorithm undergoes tripartite training with different random seeds for generalization. All experiments run on a platform with an Intel Xeon Silver 4214R CPU and NVIDIA GeForce RTX 3090 GPU.

\subsection{Experiment Results}
\subsubsection{Performance Analysis}
\begin{figure}
    \centering
    \includegraphics[width=0.9\linewidth]{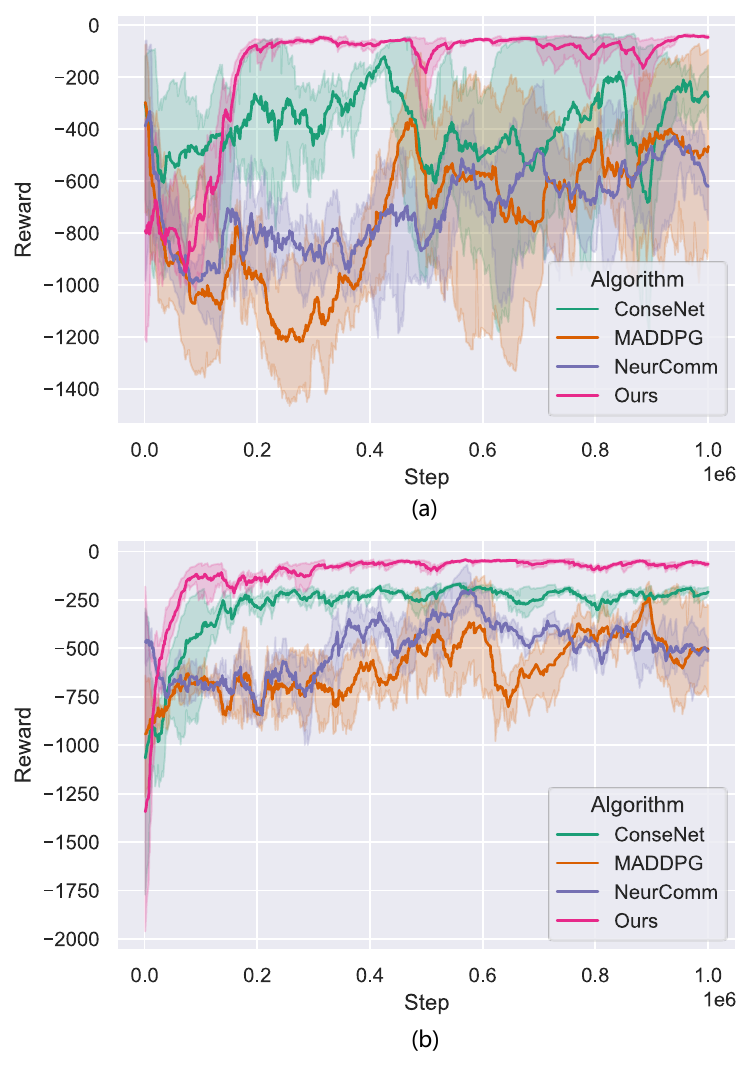}
    \caption{The average rewards of different algorithms on different scenarios, (a)Catchup scenario; (b)Slowdown scenario.}
    \label{fig:reward_curve}
\end{figure}

Fig.\ref{fig:reward_curve} compares the learning curves of our method against other baseline methods for a CACC platoon with a size of 8. ConseNet demonstrates relatively better performance among the baseline algorithms, whereas MADDPG and NeurComm struggle to converge stably in the Catchup scenario. In contrast, our method exhibits higher learning efficiency, more stable decision-making, and superior overall performance across both scenarios.

After training completion, we assess all algorithms over 50 trials under varying initial conditions, using metrics such as average headway, average velocity, and collision counts. Tab.\ref{tab:comparative_analysis} presents these evaluation results, showing our method consistently outperforms the other baselines across different scenarios. Notably, in the more challenging Slowdown scenario, MADDPG and NeurComm fail to ensure safety throughout the tests, ConseNet performs relatively better, but our method achieves the lowest collision rate.

Moreover, taking the Catchup scenario as an example, we plot the headway and velocity changes of CACC vehicles controlled by different methods. NeurComm leads to collisions around 25 seconds, causing task failure. MADDPG and ConseNet manage some level of control, yet they fall short of achieving the optimal control target within 60 seconds. Our method, however, achieves the target headway and velocity in the shortest time with the best stability.

\begin{table}[!htbp]
    \centering
    \caption{Comparative Analysis of Different Algorithms in Catchup and Slowdown Scenarios.}
    \label{tab:comparative_analysis}
    \begin{tabular}{lcccc}
        \toprule
        & ConseNet & MADDPG & NeurComm & \textbf{Ours} \\
        \midrule
        \multicolumn{5}{c}{Catchup} \\
        \midrule
        Avg Headway (m) & 27.3 & 25.7 & 27.2 & \textbf{27.7} \\
        Avg Velocity (m/s) & 15.5 & 15.3 & 16.8 & \textbf{15.1} \\
        Collision Count & \textbf{0} & 27 & 50 & \textbf{0} \\
        \midrule
        \multicolumn{5}{c}{Slowdown} \\
        \midrule
        Avg Headway (m) & 21.5 & 20.7 & \textbf{23.8} & 21.9 \\
        Avg Velocity (m/s) & 19.1 & 21.7 & 25.3 & \textbf{18.8} \\
        Collision Count & 12 & 50 & 50 & \textbf{3} \\
        \bottomrule
    \end{tabular}
\end{table}

\begin{figure}
    \centering
    \includegraphics[width=1\linewidth]{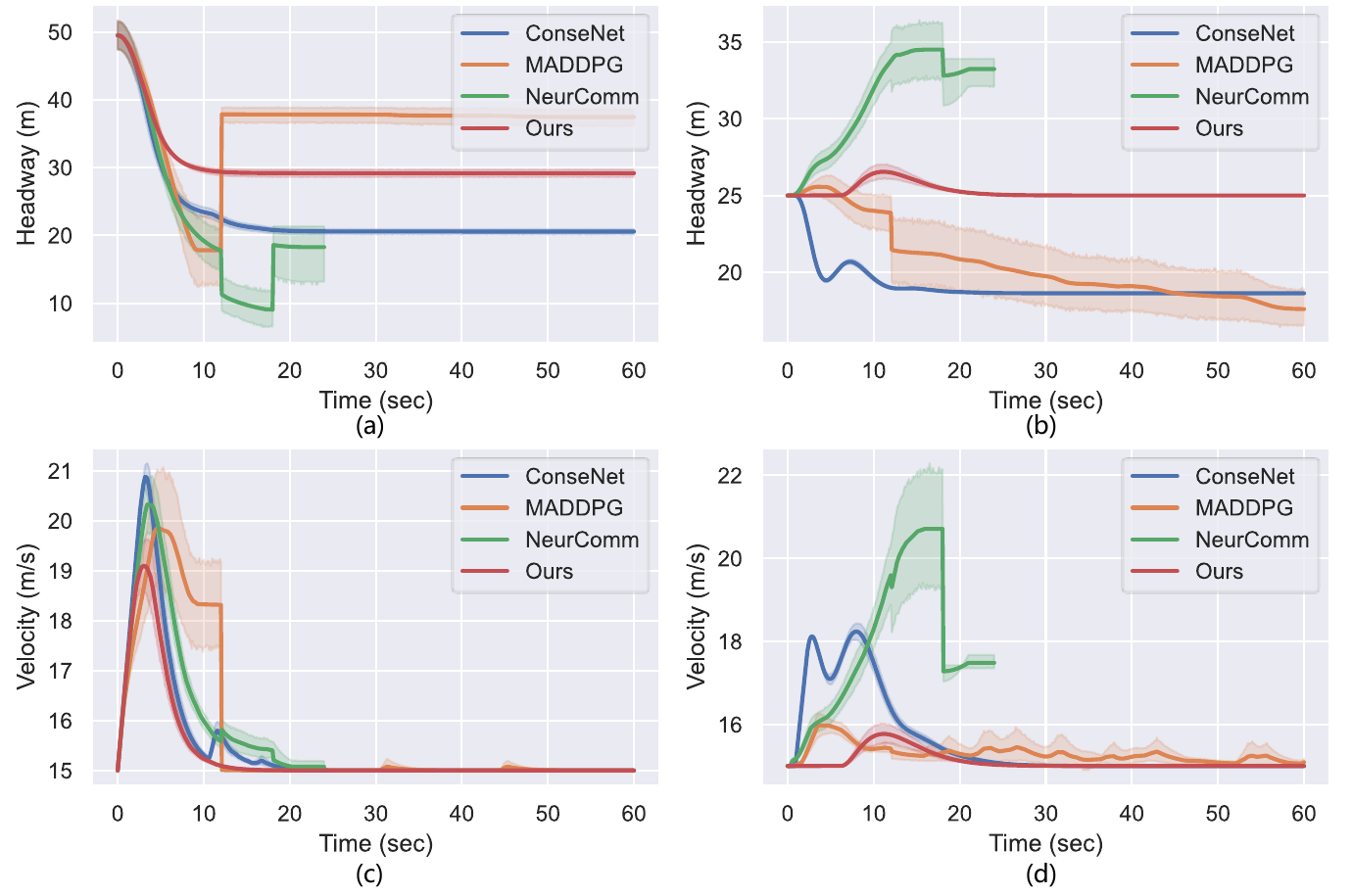}
    \caption{Headway and velocity curves in the Catchup scenario with different algorithms, (a)the headway of vehicle 1; (b)the headway of vehicle 2; (c)the velocity of vehicle 1; (d)the velocity of vehicle 2.}
    \label{fig:speed_headway_curve_2V}
\end{figure}


\subsubsection{Impact of Platoon Size}
We also investigate the impact of platoon size on our model's performance under delay considerations. 
The experiments are conducted across different platoon sizes (sizes=$\{5,8,10\}$). 
Tab. \ref{tab:avg_reward_platoon_sizes} displays the performance of various methods during the testing phase.

It is observed that with an increase in platoon size, which amplifies the difficulty and complexity of coordination among CAVs, most algorithms exhibit a performance decline in both scenarios. 
However, our algorithm consistently outperforms all other baseline methods in every scenario, with its advantage being even more pronounced in the Slowdown scenario.

\begin{table}[!htbp]
    \centering
    \caption{Average Reward of Different Algorithms with Various Platoon Sizes.}
    \label{tab:avg_reward_platoon_sizes}
    \begin{tabular}{ccccc}
        \toprule
        Platoon Size & ConseNet & MADDPG & NeurComm & \textbf{Ours} \\
        \midrule
        \multicolumn{5}{c}{Catchup} \\
        \midrule
        5 & -13.4 & -20.1 & -14.8 & \textbf{-16.0} \\
        8 & -30.9 & -636.2 & -667.8 & \textbf{-36.6} \\
        10 & -390.1 & -414.2 & -506.5 & \textbf{-43.4} \\
        \midrule
        \multicolumn{5}{c}{Slowdown} \\
        \midrule
        5 & -23.7 & -39.1 & -38.7 & \textbf{-22.8} \\
        8 & -142.7 & -488.4 & -750.8 & \textbf{-60.3} \\
        10 & -132.1 & -608.5 & -1149.6 & \textbf{-48.4} \\
        \bottomrule
    \end{tabular}
\end{table}

\section{Conclusion}
\label{sec:6}
CACC technologies are crucial for enhancing traffic efficiency and reducing energy consumption. Existing reinforcement learning-based CACC control methods have not accounted for the widespread issue of delays. To address this challenge, we proposed a delay-aware MARL framework. Leveraging a Delay-aware Markov Decision Process, we introduced a CTDE MARL framework to implement distributed control of CAVs within CACC platoons. An attention mechanism-integrated Policy network and a velocity optimization model-based action filter were designed to improve the model's performance. Our approach was experimentally validated on platoons of various sizes, demonstrating superior performance in scenarios with delays compared to baseline methods.

In the future, we aim to delve deeper into the specific types of delays within CACC platoons, such as communication and control delays, and design personalized modules to address these delays separately. Additionally, we plan to conduct experiments in more complex mixed traffic scenarios to enhance the adaptability of the model.

\bibliographystyle{IEEEtran} 
\bibliography{reference}

\vfill

\end{document}